\documentclass[runningheads]{llncs}
\usepackage[T1]{fontenc}

\usepackage{tabularx}
\usepackage{array}
\usepackage{hyperref}
\usepackage{graphicx}
\usepackage{amsmath,amssymb}
\usepackage{booktabs}
\usepackage{placeins}

\begin{document}
\title{Confidence Estimation in Automatic Short Answer Grading with LLMs}
%
%
\author{Longwei Cong\inst{1}\orcidID{0009-0008-2568-4556} \and
Sonja Hahn\inst{1}\orcidID{0000-0002-5461-6383} \and
Sebastian Gombert\inst{1}\orcidID{0000-0001-5598-9547} \and
Leon Camus\inst{1}\orcidID{0009-0008-8871-2657} \and
Hendrik Drachsler\inst{1,2}\orcidID{0000-0001-8407-5314} \and
Ulf Kroehne\inst{1, 3}\orcidID{0000-0002-0412-169X}}
%
\authorrunning{L. Cong et al.}
%
\institute{DIPF | Leibniz Institute for Research and Information in Education, 60323 Frankfurt am Main, Germany \\ 
 \email{\{l.cong,s.hahn,s.gombert,l.camus,h.drachsler,u.kroehne\}@dipf.de} \\\and
Faculty of Computer Science, Goethe University Frankfurt, 60629 Frankfurt am Main, Germany \\
  \and
 Chemnitz University of Technology, 09111 Chemnitz, Germany\\}
\maketitle              
\begin{abstract}
Automatic Short Answer Grading (ASAG) with generative large language models (LLMs) has recently demonstrated strong performance without task-specific fine-tuning, while also enabling the generation of synthetic feedback for educational assessment. Despite these advances, LLM-based grading remains imperfect, making reliable confidence estimates essential for safe and effective human-AI collaboration in educational decision-making. In this work, we investigate confidence estimation for ASAG with LLMs by jointly considering model-based confidence signals and dataset-derived uncertainty. We systematically compare three model-based confidence estimation strategies, namely verbalizing, latent, and consistency-based confidence estimation, and show that model-based confidence alone is insufficient to reliably capture uncertainty in ASAG. To address this limitation, we propose a hybrid confidence framework that integrates model-based confidence signals with an explicit estimate of dataset-derived aleatoric uncertainty. Aleatoric uncertainty is operationalized by clustering semantically embedded student responses and quantifying within-cluster heterogeneity. Our results demonstrate that the proposed hybrid confidence measure yields more reliable confidence estimates and improves selective grading performance compared to single-source approaches. Overall, this work advances confidence-aware LLM-based grading for human-in-the-loop assessment, supporting more trustworthy AI-assisted educational assessment systems.

\end{abstract}
\section{Introduction}

Automatic Short Answer Grading (ASAG) refers to the task of assessing short, natural-language student responses to objective questions using computational methods \cite{burrows2015eras}. By automating the evaluation of constructed responses, ASAG promises scalable assessment while preserving richer evidence of student understanding than multiple-choice formats \cite{livingston2009constructed}. 

Early work in ASAG primarily relied on traditional machine learning approaches, which require carefully engineered linguistic features and substantial amounts of annotated training data~\cite{magooda2016vector}. With advances in natural language processing, encoder-only pre-trained language models (PLMs) have been increasingly applied to ASAG \cite{10.1007/978-3-030-52240-7_8,gombert2023coding}. While such models achieve strong performance after task-specific fine-tuning, their application in educational settings is often impractical due to the limited availability of annotated student responses, particularly in early stages of assessment development. More recently, generative large language models (LLMs) have been explored for ASAG due to their strong language understanding and multi-step reasoning capabilities \cite{ferreira2025automatic,frohn2025automated}. Without task-specific fine-tuning, LLMs can achieve grading performance comparable to, or even exceeding, that of fine-tuned PLMs, while eliminating the need for labeled training data \cite{Cong2026ASAG}. Moreover, LLMs are capable of automatically generating feedback based on their grading decisions \cite{aggarwal2025understand}.

Despite these advances, no ASAG system achieves perfect accuracy due to linguistic variability in student responses, ambiguity in scoring criteria, and inherent disagreement among human graders, which impose an upper bound on performance \cite{burrows2015eras}. In practical assessment scenarios, a fundamental trade-off exists between grading accuracy and the human effort required for review \cite{bexte2024scoring}. Rather than fully automating grading, recent work \cite{bexte2024scoring,zehner-etal-2025-cascades} increasingly emphasizes human-in-the-loop ASAG, where automated systems assist human graders by selectively deferring uncertain responses for manual inspection. A key requirement for such human–AI collaboration is reliable confidence estimation. Existing work on confidence estimation in ASAG has largely focused on PLMs, while confidence estimation for LLMs remains insufficiently studied, despite increasing adoption in educational assessment. Moreover, existing approaches typically derive confidence solely from model-based signals, while uncertainty arising from the dataset itself remains underexplored. This limitation is particularly consequential in ASAG, where open-ended student responses are often ambiguous, underspecified, or semantically diverse \cite{burrows2015eras}. Finally, confidence estimates are often used only for binary thresholding and are rarely subjected to systematic reliability analysis, limiting their interpretability and usefulness for educators.

Our contributions are threefold. First, we provide a systematic study of confidence estimation for ASAG with LLMs by comparing verbalizing, latent, and consistency-based confidence signals. To the best of our knowledge, such an analysis has not been previously conducted in the context of ASAG. Second, we explicitly model uncertainty arising from the dataset itself by quantifying semantic heterogeneity, and propose a hybrid confidence measure that integrates model-based confidence with dataset-derived uncertainty via a probabilistic classifier. 
Third, beyond method development, we demonstrate how reliability analysis and selective prediction can be used to evaluate confidence estimates in ASAG, and show that the proposed hybrid confidence scores are better aligned with empirical grading accuracy, making them more suitable for selective grading scenarios in human-in-the-loop assessment.

\section{Background}

\subsection{Confidence Estimation in Automatic Short Answer Grading}

In practical educational assessment settings, automated grading systems are commonly integrated into human-in-the-loop workflows, where automated scores serve as decision support for human graders in detecting and correcting potential grading errors \cite{attali2006automated}.

Several studies have investigated selective grading frameworks that combine automated scoring with human review. Funayama et al. \cite{funayama2022balancing} proposed an approach in which ASAG systems share the grading workload with human raters by estimating a confidence score for each prediction. Responses with confidence below a predefined threshold are deferred to human graders, where the threshold is calibrated on a development dataset to constrain automated scoring errors. Confidence is estimated using different approaches, including model-based posterior probabilities, distance-based metrics such as Trust Score \cite{jiang2018trust}, and predictive uncertainty estimates obtained from regression models. Similarly, Bexte et al. \cite{bexte2024scoring} employed a similarity-based classification approach for ASAG, where the similarity score underlying the classification decision is directly used as a confidence estimate. In their work, confidence thresholds are predefined based on classification performance metrics. More recently, Zehner et al. \cite{zehner-etal-2025-cascades} introduced a semi-automatic grading framework that cascades fine-tuned PLMs and human graders. The confidence threshold determining whether a response is graded automatically or forwarded to human raters is selected by maximizing Youden’s index, which quantifies the trade-off between sensitivity and specificity of classification. 

While these studies demonstrate the value of confidence-based selective grading for ASAG, they primarily focus on PLMs. In contrast, confidence estimation for ASAG with LLMs remains largely unexplored.

\subsection{Confidence Estimation with LLMs}

LLMs have demonstrated remarkable performance across a wide range of tasks and domains. However, their outputs can be unreliable due to factual errors and hallucinations \cite{geng2024survey}. Consequently, reliable confidence estimation is essential for enabling safe and effective human–AI collaboration, allowing systems to defer uncertain predictions to human experts. Broadly, existing approaches to confidence estimation for LLMs can be categorized into four classes \cite{shorinwa2025survey}. 

Verbalizing-based methods explicitly prompt the model to generate both a task prediction and a self-reported confidence score \cite{tian2023just}. In contrast, latent-based methods derive confidence from internal model signals, such as token-level probability distributions, which are aggregated using task-specific metrics \cite{kadavath2022languagemodelsmostlyknow}. Consistency-based methods estimate confidence by measuring the stability of model predictions across multiple sampled outputs, obtained by varying decoding parameters (e.g., temperature) or by perturbing the input through paraphrasing \cite{xiong2023can}. Finally, mechanistic interpretability \cite{bereska2024mechanistic} has been proposed as a direction for understanding uncertainty in LLMs, although it has not yet been widely applied
in practice.

Despite their methodological differences, the above approaches estimate confidence primarily from model-based signals. In ASAG tasks, however, uncertainty may also arise from the dataset itself. Such dataset-derived uncertainty (i.e., aleatoric uncertainty) is not explicitly modeled in existing confidence estimation methods for LLM-based grading.

\subsection{Uncertainty Sources in LLMs}

There are two primary sources of uncertainty in machine learning, namely aleatoric uncertainty and epistemic uncertainty \cite{hullermeier2021aleatoric}. In the context of LLMs, aleatoric uncertainty arises from ambiguous or incomplete information as well as from inherent properties of natural language, and is therefore irreducible by improving the model alone \cite{xia-etal-2025-survey}. Such uncertainty is particularly prevalent in ASAG, where student responses are often brief, underspecified, or linguistically ambiguous \cite{haller2022survey}. Prior works \cite{ghandeharioun2019characterizing,uma2021learning} show a strong correlation between aleatoric uncertainty and human annotator disagreement, suggesting that data-related uncertainty can be approximated through patterns of disagreement in the responses. In contrast, epistemic uncertainty arises from insufficient knowledge or limited training data, reflecting the limitation of the grading model itself \cite{xia-etal-2025-survey}. In LLMs-based systems, epistemic uncertainty is commonly estimated using model-based signals as described above.



\section{Method}

Motivated by the relationship between ambiguity and annotator disagreement, we explicitly model aleatoric uncertainty via label heterogeneity in a semantic embedding space, where higher heterogeneity indicates greater ambiguity in the underlying student answers. To estimate epistemic uncertainty, we derive confidence signals directly from the LLMs using three complementary approaches: verbalizing, latent, and consistency-based methods across multiple generations. Finally, since these uncertainty sources estimate fundamentally different aspects of unreliability, we combine them by training a probabilistic classifier with calibration to produce a hybrid confidence score. 

\subsection{Confidence Estimation from LLMs}
\label{method:confidence}

We use the open-weight large language model \emph{gpt-oss-20b} \cite{agarwal2025gpt}, which represents a state-of-the-art model for ASAG \cite{Cong2026ASAG}. Following \cite{farquhar2024detecting}, we first set the decoding temperature to 0.1 and generate a single grading decision as the final predicted label $\hat{y}$. 

To obtain verbalizing-based confidence, we explicitly instruct the model in the prompt to provide a confidence score for its judgment as a probability in the range $[0,1]$, following the prompting protocol proposed by \cite{tian2023just}. The prompts are provided in the \href{https://anonymous.4open.science/api/repo/appendix_aied26-BE26/file/appendix.pdf}{Digital Appendix}. 

For latent-based confidence, following \cite{kadavath2022languagemodelsmostlyknow}, we compute the conditional log-likelihood \(\log p(y \mid u)\) for each label \(y \in \mathcal{Y}\), where \(u\) is the input prompt prefix and \(y\) contains only the label token(s), without any reasoning tokens. These label-wise log-likelihoods are normalized across all candidates using a softmax function. The latent confidence is defined as the normalized score assigned to the finial predicted label:
\begin{equation}
s^{\text{lat}}
= \frac{\exp\!\left(\log p(\hat{y} \mid u)\right)}
{\sum_{y \in \mathcal{Y}} \exp\!\left(\log p(y \mid u)\right)}=\frac{ p(\hat{y} \mid u)}
{\sum_{y \in \mathcal{Y}} p(y \mid u)}.
\end{equation}

For consistency-based confidence, the model is sampled multiple times at temperatures 0.2, 0.4, 0.6, 0.8, and 1.0, yielding a set of predicted labels \(\{ \hat{y}_1, \ldots, \hat{y}_N \}\). Following \cite{xiong2023can}, the consistency confidence for the finial predicted label is defined as
\begin{equation}
s^{\text{cons}} = \frac{1}{N} \sum_{i=1}^{N} \mathbb{I}(\hat{y}_i = \hat{y}),
\end{equation}
where \(\mathbb{I}(\cdot)\) denotes the indicator function.

Given the binary scoring setting, all confidence scores are oriented to represent the model’s confidence that a student response is correct. 
Let $s_i' \in [0,1]$ denote a raw confidence signal. If $\hat{y}_i = 1$, we set $s_i = s_i'$; otherwise, we set $s_i = 1 - s_i'$. This transformation ensures a consistent interpretation of confidence across predictions and facilitates subsequent calibration and confidence fusion.


\subsection{Uncertainty Estimation from Dataset}

To estimate dataset-derived uncertainty, we adopt an empirical measure of aleatoric uncertainty that captures ambiguity in student responses through semantic heterogeneity. Let $\mathcal{D} = \{(x_i, y_i)\}$ denote a set of student responses $x_i$ without explicit question context, where $y_i \in \mathcal{Y}$ is the corresponding label indicating whether the response is correct or incorrect. We split the data into two stratified subsets, a calibration subset $\mathcal{D}_{\mathrm{cal}}$ and a test subset $\mathcal{D}_{\mathrm{test}}$. The calibration subset is used to estimate semantic heterogeneity and to learn the classifier’s calibration mapping.

To represent semantic similarity between student responses, each response $x_i$ is embedded into a semantic vector space using a pretrained sentence-transformers model \emph{all-MiniLM-L6-v2} \cite{10.5555/3495724.3496209}, following prior work \cite{bexte2024scoring}.
We then apply agglomerative hierarchical clustering with Ward linkage \cite{ward1963hierarchical} to the embeddings of the calibration subset. This results in a partition of $\mathcal{D}_{\mathrm{cal}}$ into $K$ clusters, with each response assigned to a cluster index $c_i \in \{1,\dots,K\}$. Ward linkage is chosen because it minimizes within-cluster variance and produces compact clusters of semantically similar responses \cite{petukhova2025text,zehner2016automatic}.

For each cluster index $k \in \{1,\dots,K\}$, we estimate the label distribution
\begin{equation}
p(y=j \mid c=k) = \frac{1}{|S_k|} \sum_{i \in S_k} \mathbb{I}(y_i = j), \quad j \in \mathcal{Y},
\end{equation}
where $S_k = \{i \mid c_i = k\}$ is the set of calibration responses in cluster $k$, and $\mathbb{I}(\cdot)$ is the indicator function. We quantify the degree of heterogeneity within each cluster using the normalized Shannon entropy
\begin{equation}
H_k = - \frac{1}{\log |\mathcal{Y}| }\sum_{j \in \mathcal{Y}} p(y=j \mid c=k)\log p(y=j \mid c=k),
\end{equation}
where $H_k$ reflects the degree of semantic heterogeneity associated with cluster $k$. Low entropy values indicate that semantically similar responses are consistently assigned the same score, while higher values indicate increased ambiguity and disagreement.

Each calibration response inherits the aleatoric uncertainty of its cluster, $s^{\mathrm{alea}}_i = H_{c_i}$. To assign aleatoric uncertainty to unseen test responses, we compute the centroid of the embeddings corresponding to the calibration responses for each cluster. For a test response, we identify the nearest cluster centroid using Euclidean distance and assign the corresponding cluster entropy as its aleatoric uncertainty.


\subsection{Confidence Fusion}

The confidence and uncertainty estimation methods described above capture complementary aspects of uncertainty. Model-based confidence scores reflect epistemic uncertainty of the LLMs, whereas the semantic heterogeneity reflects response ambiguity and thus provides an aleatoric uncertainty proxy. Since these sources are not directly comparable and may interact nonlinearly, we fuse them into a single hybrid confidence score via supervised calibration.


For each response, we construct a feature vector
\[
\mathbf{z}_i =
\big[
s^{\mathrm{verb}}_i,\;
s^{\mathrm{lat}}_i,\;
s^{\mathrm{cons}}_i,\;
s^{\mathrm{alea}}_i,\;
\ell_i\;
\big],
\]
where $s^{\mathrm{verb}}_i$, $s^{\mathrm{cons}}_i$, and $s^{\mathrm{lat}}_i$ denote the oriented verbalizing, latent, and consistency-based confidence scores derived from the LLMs, respectively, each oriented as described in Section~\ref{method:confidence} so that higher values indicate greater confidence that the response is correct; $s^{\mathrm{alea}}_i$ represents the proxy of aleatoric uncertainty; and $\ell_i$ denotes the token length of the student response. We include these length features because response verbosity and generated rationale length can correlate with both ambiguity and model behavior \cite{horbach2019influence}.

We train a probabilistic classifier to map $\mathbf{z}_i$ to the probability that a student response is correct ($y_i = 1$). 
Concretely, we learn a function $g(\cdot)$ such that
\begin{equation}
p_i = g(\mathbf{z}_i) \approx \mathbb{P}(y_i = 1 \mid \mathbf{z}_i).
\end{equation}
In our implementation, $g(\cdot)$ is a Random Forest classifier with 500 trees, trained using the human labels $y_i$ as supervision. This model is chosen for its ability to model nonlinear feature interactions and its robustness to heterogeneous input features. Because raw ensemble probabilities are often not perfectly calibrated, we apply post-hoc probability calibration using Platt scaling \cite{silva2023classifier} with five-fold cross-validation. This yields a calibrated hybrid confidence score $p_i \in [0,1]$. For a fair comparison, we apply the same calibration procedure to each model-based confidence signal in the evaluation.

To isolate the contribution of uncertainty that we calculated from the semantic heterogeneity, we train two fusion variants: a \emph{hybrid model with aleatoric features} (including $s^{\mathrm{alea}}_i$) and a \emph{hybrid model without aleatoric features}, while keeping all other inputs identical.

\subsection{Evaluation Methodology}

\label{sec:evaluation}

We evaluate the quality of confidence estimates using two criteria: \emph{selective prediction performance} and \emph{reliability}. Our evaluation focuses on how confidence can be used to selectively accept reliable predictions and provide reliable confidence estimates.

\subsubsection{Selective Prediction Analysis}

Selective prediction performance assesses a method’s ability to discriminate between correct and incorrect grading decisions based on estimated confidence. Two commonly used metrics for this purpose are the receiver operating characteristic (ROC) curve \cite{bradley1997use} and the accuracy--rejection curve (ARC) \cite{nadeem2009accuracy}, along with their corresponding summary measures, area under the ROC curve (AUROC) and area under the ARC (AUARC).

The ROC curve evaluates how well a confidence score separates correct from incorrect predictions across all possible thresholds by plotting the true positive rate against the false positive rate. AUROC therefore reflects the overall ranking quality of a confidence measure, independent of any specific operating point. In contrast, the ARC captures the trade-off between prediction accuracy and human effort reduction by plotting the accuracy of retained predictions as a function of the rejection rate, reflecting selective grading behavior under increasing rejection of low-confidence predictions.
The AUARC summarizes this behavior across rejection levels.




\subsubsection{Reliability Analysis}

In addition to selective prediction, we assess whether confidence scores are reliable, i.e., whether predicted probabilities correspond to empirical correctness frequencies. A reliable confidence score ensures that, among all predictions assigned a confidence of $p$, approximately $p$ fraction are correct.

We evaluate calibration using reliability diagrams, which compare predicted probabilities with observed accuracies across confidence bins. Predictions are grouped into a fixed number of bins based on their confidence scores, and for each bin we compute the mean predicted confidence and the empirical accuracy. To quantify calibration quality, we report three standard metrics. The \emph{Brier score} measures the mean squared error between predicted probabilities and binary correctness labels. The \emph{Expected Calibration Error} (ECE) computes the weighted average absolute difference between predicted confidence and empirical accuracy across bins, while the \emph{Maximum Calibration Error} (MCE) reports the largest such deviation. 

\section{Experiments}

\subsection{Dataset}

We evaluate our approach on \emph{SciEntsBank} \cite{dzikovska2013semeval}, a widely used benchmark dataset for ASAG containing nearly 11,000 student responses to 197 science questions across 15 domains. The dataset provides human-annotated labels reflecting the semantic correctness of student answers. We focus on the \texttt{Test\_UD} split, treating correct responses as positive instances and grouping all remaining labels as incorrect. This split includes 4,562 responses, with 1,917 correct and 2,645 incorrect answers, covering diverse error types such as contradictory, partially correct, and irrelevant response. The substantial heterogeneity among incorrect answers makes this split particularly suitable for studying confidence estimation. For confidence calibration and dataset-derived uncertainty estimation, we further divide the evaluation split into a calibration subset (10\%) and a held-out test subset (90\%) using stratified sampling, so that the calibration data matches the distribution of the held-out test data.


\subsection{Results}

\subsubsection{Selective Prediction Analysis}

\begin{figure}
\centering
\includegraphics[width=\linewidth]{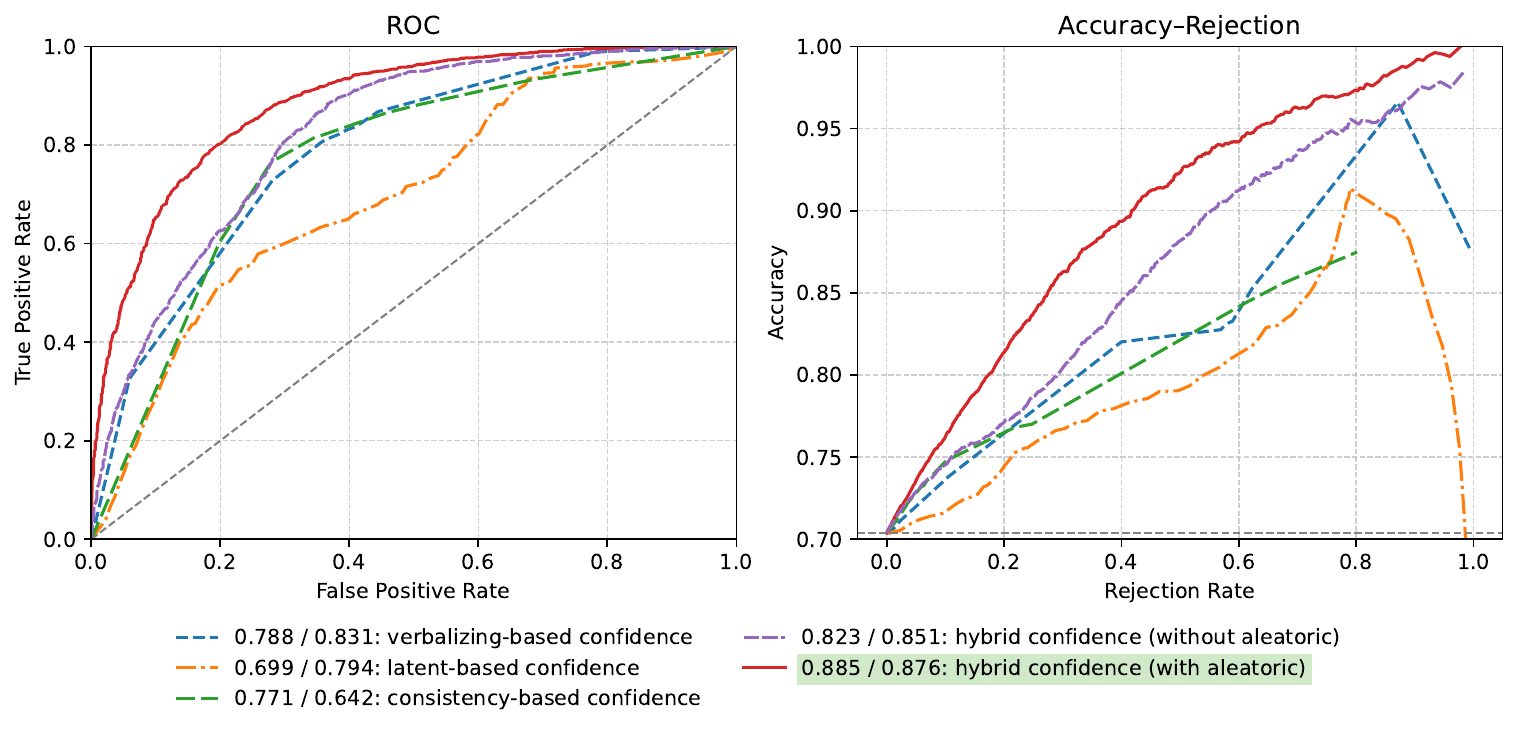}
\caption{ROC (left) and ARC  (right). The accuracy is 0.704 without any confidence-based selection. The legend reports AUROC as the first value and AUARC as the second.}
\label{plt:aoc_arc}
\end{figure}

Fig. \ref{plt:aoc_arc} evaluate the effectiveness of different confidence estimation strategies for selective prediction. The ROC analysis shows that all confidence signals perform better than random guessing, but with substantial variation in discrimination ability. Among the single-source methods, verbalizing and consistency-based confidence achieve moderate performance, whereas latent confidence yields the lowest AUROC, indicating that raw token-level probability alone is insufficient to reliably distinguish correct from incorrect responses. This limitation is further reflected in the ARC analysis, where latent confidence leads to only marginal accuracy improvements and degrades sharply at higher rejection rates.

In contrast, the hybrid confidence models consistently dominate both evaluations. The hybrid model incorporating aleatoric uncertainty achieves the highest AUROC and yields the strongest accuracy gains as low-confidence responses are progressively rejected. For example, at a rejection rate of 0.4, corresponding to filtering out 40\% of low-confidence predictions, the accuracy of the remaining responses increases to approximately 0.900, compared to 0.704 when no confidence-based selection is applied. The hybrid model without aleatoric uncertainty remains competitive but exhibits consistently lower discrimination and slower accuracy improvements, underscoring the added value of explicitly modeling dataset-derived uncertainty via semantic heterogeneity.


\subsubsection{Reliability Analysis}
\begin{figure}
\centering
\includegraphics[width=\linewidth]{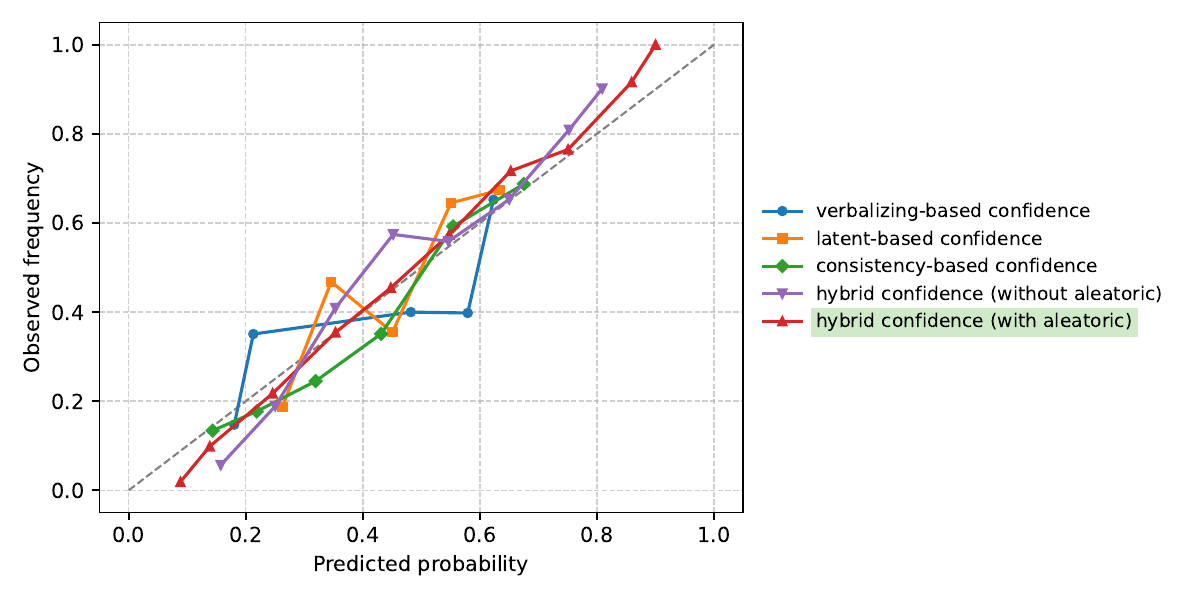}
\caption{Reliability diagrams for each confidence estimation method. The diagonal line represents perfect calibration, where predicted confidence matches the observed empirical accuracy.}
\label{plt:reliability}
\end{figure}

Fig. \ref{plt:reliability} presents reliability diagrams for the different confidence estimation methods, while Table \ref{tab:calibration_metrics} reports quantitative calibration metrics, including the Brier score, expected calibration error (ECE), and maximum calibration error (MCE). 

Among the single-source methods, verbalizing-based confidence shows clear deviations from the diagonal, particularly in the mid-confidence range, where predicted confidence exceeds empirical accuracy, indicating overconfidence. This behavior is also reflected in its relatively high MCE. Latent-based confidence exhibits even inferior calibration, achieving the highest Brier score and ECE. Consistency-based confidence follows the diagonal more closely and attains the lowest ECE, suggesting good average calibration. However, its relatively high MCE indicates that substantial calibration errors still occur in certain confidence regions.

The hybrid confidence models yield the most reliable calibration overall. In particular, the hybrid model incorporating aleatoric uncertainty closely tracks the diagonal across confidence bins and achieves the best Brier score and lowest MCE, indicating both accurate probability estimates and reduced worst-case miscalibration. Notably, this model shows improved calibration in the mid-confidence range, where dataset-derived ambiguity is most pronounced. 
The hybrid model without aleatoric uncertainty also improves calibration compared to single-source methods, but exhibits consistently larger deviations.


\begin{table}
\centering
\caption{Calibration performance of different confidence estimation methods.}
\label{tab:calibration_metrics}
\begin{tabular}{lcccc}
\toprule
Method  & Brier $\downarrow$ & ECE $\downarrow$ & MCE $\downarrow$ \\
\midrule
verbalizing-based confidence & 0.194 & 0.051 & 0.259 \\
latent-based confidence    & 0.218 & 0.096 & 0.221 \\
consistency-based confidence   & 0.186 & \textbf{0.029} & 0.279 \\
hybrid confidence (without aleatoric)   & 0.172 & 0.066 & 0.155 \\
hybrid confidence (with aleatoric)   & \textbf{0.138} & 0.044 & \textbf{0.100} \\
\bottomrule
\end{tabular}
\end{table}

\section{Discussion}

\subsection{Responses to Results}

The ROC curve and ARC show that the proposed hybrid confidence measure consistently outperforms all single-source confidence estimates. For a fixed rejection level, the hybrid approach achieves higher grading accuracy, while for a fixed accuracy level, it reduces more human grading effort. This demonstrates that incorporating dataset-derived uncertainty is particularly effective for identifying subsets of responses that can be graded with high confidence. From a human--AI collaboration perspective, ARC provides a practical decision-support tool for selective delegation,
supporting the selection of operating points that balance grading accuracy against the amount of manual review required. Analogous to the Youden index \cite{youden1950index} used in ROC analysis, an optimal trade-off point can be defined by jointly maximizing accuracy and human effort reduction. This provides a principled basis for selective grading decisions, rather than relying on threshold selection strategies solely based on predefined error metrics \cite{bexte2024scoring,funayama2022balancing,zehner-etal-2025-cascades}.

The reliability diagrams further reveal substantial differences among confidence estimation strategies. Even after calibration, verbalizing, latent, and consistency-based confidence signals exhibit noticeable miscalibration.
Latent confidence performs worst, which is consistent with findings reported in \cite{xia-etal-2025-survey}. This behavior can be explained by the generative nature of LLMs. Unlike discriminative classifiers such as PLMs, LLMs produce grading decisions as token sequences \cite{zhao2023survey}, and latent confidence is typically derived from the probability of the final label token \cite{xia-etal-2025-survey}. However, this probability is computed after the model has conditioned on the input and committed to a specific decision, and therefore primarily reflects decoding determinism rather than true classification uncertainty. While consistency-based confidence achieves low average calibration error, its higher maximum calibration error indicates the presence of localized failure regions, which are particularly problematic in educational settings where rare but severe miscalibration can undermine \cite{shorinwa2025survey}. In contrast, the hybrid confidence measure that incorporates aleatoric uncertainty shows markedly improved calibration, with predicted confidence closely matching empirical accuracy. This enables the reliable prioritization of student responses into review queues, where high-confidence responses can be graded automatically, while low-confidence cases are flagged at different risk levels for human review. 

This finding suggests that model-based confidence alone is insufficient to adequately capture uncertainty in ambiguous student responses. The improvements achieved by the hybrid confidence measure can be attributed to explicitly modeling aleatoric uncertainty using semantic heterogeneity as a proxy. Importantly, aleatoric uncertainty in ASAG often reflects genuine pedagogical ambiguity, such as partially correct reasoning or underspecified explanations, rather than model deficiencies \cite{hullermeier2021aleatoric}. Accounting for this form of uncertainty allows confidence estimates to better align with the inherent indeterminacy of open-ended student responses. While advances in LLM reasoning capabilities \cite{zhao2023survey} or the use of LLM-based multi-agent systems \cite{wei2026agentic} may mitigate epistemic uncertainty, aleatoric uncertainty remains intrinsic to open-ended assessment tasks. Consequently, explicitly estimating aleatoric uncertainty is essential for reliable confidence estimation and safe deployment of ASAG systems in human–AI collaborative settings.

\subsection{Limitation and Future Directions}

Future work could evaluate the proposed methods on a broader range of LLMs and datasets. Due to hardware limitations, this study focuses on a state-of-the-art open-weight model that can be hosted on the available infrastructure. Extending the analysis to models with different architectures and sizes, as well as to datasets from diverse domains and annotation schemes, would help assess the generalizability of our findings.

Moreover, alternative proxies for estimating aleatoric uncertainty remain to be explored. For example, psychometric models such as Item Response Theory (IRT) \cite{lord2012applications} could be used to capture item-level structural uncertainty, providing complementary signals to semantic heterogeneity.

Future work could explore integrating confidence-aware ASAG systems into real-world grading workflows, including user studies with educators to understand how confidence estimates are interpreted and used.
\section{Conclusion}

In this work, we studied confidence estimation for ASAG with LLMs, considering both model-based confidence signals and dataset-derived uncertainty. By comparing verbalizing, latent, and consistency-based confidence strategies, we showed that model-based confidence alone is insufficient to reliably capture uncertainty in ASAG. We further proposed a hybrid confidence framework that integrates dataset-derived aleatoric uncertainty, estimated via semantic heterogeneity, with model-based confidence signals, resulting in improved selective grading and reliability performance. From a human–AI collaboration perspective, our results indicate that accuracy–rejection curves, together with reliability analysis, provide a principled evaluation perspective for interpreting calibrated confidence scores in selective grading scenarios. In particular, the hybrid confidence measure yields confidence estimates that are better aligned with empirical grading accuracy, offering educators an interpretable reference for prioritizing automated grading while identifying responses that may benefit from human review.
%
%
%
\FloatBarrier

\begin{credits}
\subsubsection{\ackname} This research was conducted within the project “Assessment for Learning with AI (ALwAI)” funded by the Leibniz Association under the Leibniz Competition (project no. T163/2024).
\end{credits}

\bibliographystyle{splncs04}
\bibliography{07references}

%
\end{document}